\documentclass[conference]{IEEEtran}
\IEEEoverridecommandlockouts
\usepackage{cite}
\usepackage{amsmath,amssymb,amsfonts}
\usepackage{algorithmic}
\usepackage{graphicx}
\usepackage{textcomp}
\usepackage{xcolor}
\usepackage{array}
\usepackage{booktabs}
\begin{document}
\bstctlcite{IEEEexample:BSTcontrol}

\title{Unsupervised Hebbian Learning on Point Sets in StarCraft II\\
\thanks{This work was supported by Office of Naval Research under Grant Number N000142012432. The views and conclusions contained in this document are those of the authors and should not be interpreted as representing the official policies, either expressed or implied, of the Office of Naval Research or the U.S. Government}
}

\author{
\IEEEauthorblockN{Beomseok Kang, Harshit Kumar, Saurabh Dash, Saibal Mukhopadhyay}
\IEEEauthorblockA{\textit{School of Electrical and Computer Engineering} \\
\textit{Georgia Institute of Technology}\\
Atlanta, GA, USA\\
\{beomseok, hkumar64, saurabhdash, smukhopadhyay6\}@gatech.edu}
}

\maketitle

\begin{abstract}
Learning the evolution of real-time strategy (RTS) game is a challenging problem in artificial intelligent (AI) system. In this paper, we present a novel Hebbian learning method to extract the global feature of point sets in StarCraft II game units, and its application to predict the movement of the points. Our model includes encoder, LSTM, and decoder, and we train the encoder with the unsupervised learning method. We introduce the concept of neuron activity aware learning combined with k-Winner-Takes-All. The optimal value of neuron activity is mathematically derived, and experiments support the effectiveness of the concept over the downstream task. Our Hebbian learning rule benefits the prediction with lower loss compared to self-supervised learning. Also, our model significantly saves the computational cost such as activations and FLOPs compared to a frame-based approach.

\end{abstract}


\section{Introduction}
Games are frequently used platforms to evaluate artificial intelligent (AI) systems. Atari \cite{Atari}, Chess \cite{Chess}, and Go \cite{Go} games are well known examples where game AI demonstrates successful progress. Recent attention on game AI is lying in real-time strategy (RTS) games as these games possess more complex setting and situation. StarCraft II is one of the most popular and challenging RTS game not only for users but also for AI researchers. There exist numerous different strategies, and more importantly, players continually change their tactical short-term and long-term planning. This non-deterministic nature of the game makes learning the game evolution demanding.

\begin{figure}
\centerline{\includegraphics[width=\columnwidth]{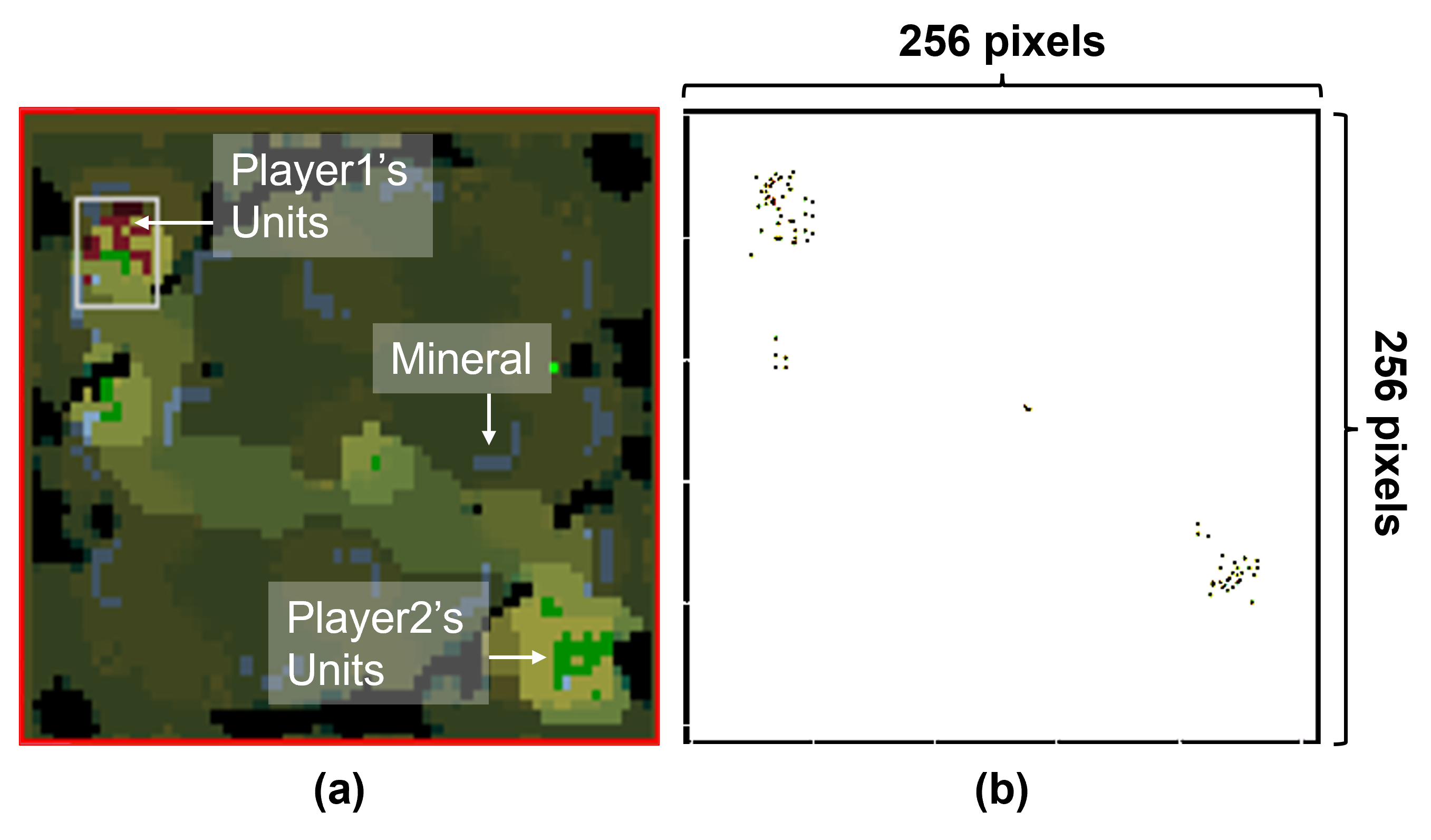}}
\caption{(a) shows the game map given to two players. Green and red pixels are the game units of two players, and blue pixels are resources (mineral). Other pixels indicate empty terrain. The units from two players encounter in player1's territory, and other player2's units are heading to the area. (b) shows the corresponding map only with the units in higher resolution. Dark pixels indicate the presence of the units, and most of other pixels are empty.}
\label{figure_Replay}
\vspace{-10pt}
\end{figure}

Most of existing works have focused on the control of the game agents based on reinforcement learning \cite{GrandMasterSC2, PySC2}. Recently, a deep neural network based approach has been studied to understand the hidden information in the game \cite{DefogGan}. It processes the map in the game as input with convolutional neural networks \cite{PySC2, DefogGan}. However, the game map is very sparse compared to the density of game units. For example, Fig. \ref{figure_Replay}(a) is the game map given to two players during a match, and the majority of areas are not in use. This is because each of players is limited to generate a maximum 200 number of units. Fig. \ref{figure_Replay}(b) shows that the sparsity of units is extremely high if we assume the resolution of the map is (256, 256) so that each unit is represented by a pixel. In this case, maximum 400 units from two players account for only 0.6\% of entire pixels. It motivates that directly processing the units as a point set will be a more efficient and effective way to understand the game evolution. However, only few works have been reported for learning the dynamics of a point set in StarCraft II by supervised learning \cite{TemporalPoint}.

In this paper, we present a completely unsupervised learning method to represent a point set in StarCraft II. We mainly develop Hebbian learning and k-Winner-Takes-All (k-WTA) algorithms to extract the global feature of a point set. Hebbian learning rule is inspired from the biological mechanism of synaptic plasticity. As weight update by Hebbian learning is based on the activity of pre-synaptic and post-synaptic neurons, no supervision or backpropagation is required. However, most of prior works on Hebbian learning is limited to multilayer perceptron or convolutional neural networks with MNIST or CIFAR-10 dataset \cite{SampleEfficientHebbian, CompetingHebbian, HebbNet, MultiLayerHebbian, HebbianCluster}. Our Hebbian learning rule is developed with focus on a point set representation, and we also introduce neuron activity aware learning combined with k-WTA.

We design three modules including encoder, LSTM, and decoder to predict the movement of units in the future. Hebbian learning is used for training the encoder to generate the fixed-dimension latent vector from point set with variable size. Long short-term memory (LSTM) and decoder are combined with the encoder, and they are trained in latent space by self-supervised learning. We mainly investigate whether the latent vector trained by Hebbian learning is a useful form to learn the dynamics of a point set. We observe the encoder trained by our unsupervised learning method outperforms in terms of the prediction loss compared to end-to-end self-supervised learning. This paper makes the following key contributions:
\begin{itemize}
\item We develop a novel Hebbian learning and k-Winner-Takes-All based unsupervised learning method for a point set in StarCraft II. The unsupervised learning benefits the feature extraction of a point set for the reconstruction and prediction of a game map.
\item We introduce the concept of neuron activity in Hebbian learning rule combining with k-WTA. We also mathematically derive the optimal neuron activity, and experimentally support the effectiveness of the concept.
\item Our unsupervised learning method achieves a lower prediction loss than self-supervised learning in the same model. Also, compared to a frame-based approach, our model shows the \(\times110\) less activations and \(\times793\) less FLOPs with the lower distance error.
\end{itemize}

\section{Proposed Approach}
\subsection{Encoder Architecture}
\begin{figure}
\centerline{\includegraphics[width=\columnwidth]{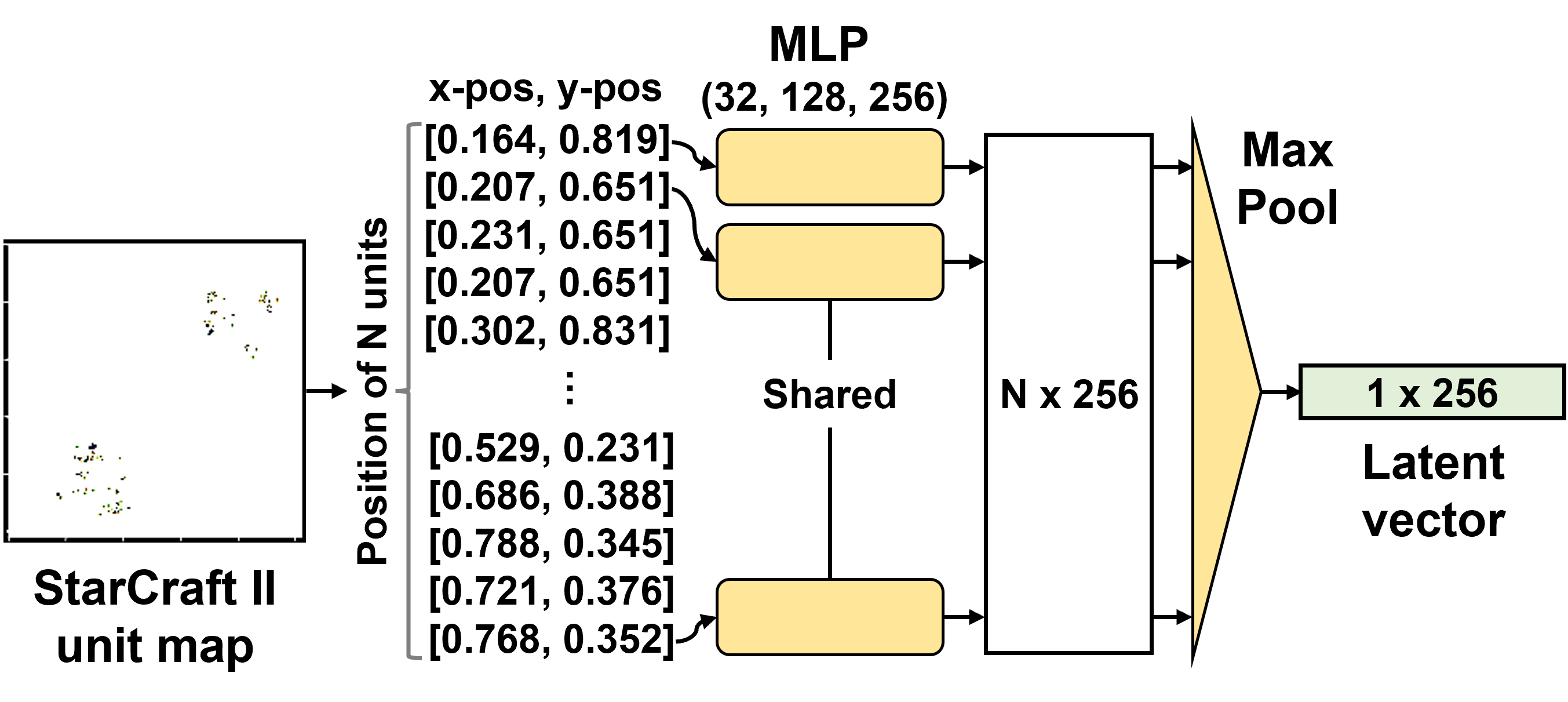}}
\caption{schematic of the encoder architecture where Hebbian learning is used to update weight parameters. The positional information of units in (N, 2) shape input is transformed to a fixed-dimension (1, 256) latent vector.}
\label{figure_Encoder_Architecture}
\end{figure}

We encode the point set from a StarCraft II unit map to a fixed-dimension latent vector. As the number of units changes during the playing, the set to vector encoding can simplify the design of following models. Our encoder is motivated from PointNet \cite{PointNet}, and Fig. \ref{figure_Encoder_Architecture} shows the schematic of the encoder architecture. The encoder extracts the global feature of a point set by combining multilayer perceptron (MLP) and MaxPooling layer. We take advantages of the simple architecture with the conventional perceptron model so that Hebbian learning can be still applied to process a point set. Between the three layers of MLP, ReLU is used as an activation function, and the output vector of each unit is normalized by their maximum value. Also, the normalized output in each layer is filtered by k-WTA.

\subsection{Hebbian Learning on Point Sets}
Equation (\ref{equation_instar_rule}) is a variant of Hebbian learning rule named Grossberg's instar rule \cite{InstarRule}.
\begin{equation}
\Delta{\textbf{w}} \propto y(\textbf{x}-\textbf{w})
\label{equation_instar_rule}
\end{equation}
where \(\textbf{x}\) and \(\textbf{w}\) are an input vector and weight vector, and \(y\) is the related output. Our Hebbian learning rule is based on the instar rule, and it inherently incorporates the distance term \((\textbf{x}-\textbf{w})\). As the direction of the weight update is from the weight vector to the input, the weight vectors are converged to the points after training \cite{SampleEfficientHebbian}.
\begin{equation}
\Delta\textbf{w}_{j} = \eta\frac{1}{N}\sum_{i}{f(|\textbf{x}_{i}-\textbf{w}_{j}|)(\textbf{x}_{i}-\textbf{w}_{j})}
\label{equation_learning_rule}
\end{equation}
\begin{equation}
  f(|\textbf{x}_{i}-\textbf{w}_{j}|) =  
    \begin{cases}
      1 & \text{if} \: i = \underset{k}{\mathrm{arg\:min}}\:{|\textbf{x}_{k}-\textbf{w}_{j}|} \\
      0 & \text{otherwise}
    \end{cases}       
\label{equation_distance_training}
\end{equation}

We modify the instar rule to incorporate \(f(|\textbf{x}_{i}-\textbf{w}_{j}|)\) instead of the output \(y\) in equation (\ref{equation_learning_rule}). \(\eta\) is learning rate, and we set 0.01 for all experiments in this paper. \(N\) is the number of points in a map, hence we calculate the average weight update. Equation (\ref{equation_distance_training}) describes that the output is 1 only if the weight vector (\(\textbf{w}_{j}\)) is closest with the point (\(\textbf{x}_{i}\)). With the consideration of the Euclidean distance between \(\textbf{x}\) and \(\textbf{w}\), neurons competitively learn different spatial points. In other words, the binary output of the function \(f(|\textbf{x}_{i}-\textbf{w}_{j}|)\) determines whether the weight of the neuron is to be updated. Though the output \(y\) is not directly used in the learning rule, the distance measure indirectly uses the output in the last term of \(|\textbf{x}_{i}-\textbf{w}_{j}|^{2}=|\textbf{x}_{i}|^{2}+|\textbf{w}_{j}|^{2}-2\textbf{x}_{i}\cdot\textbf{w}_{j}\). 

\begin{figure}
\centerline{\includegraphics[width=\columnwidth]{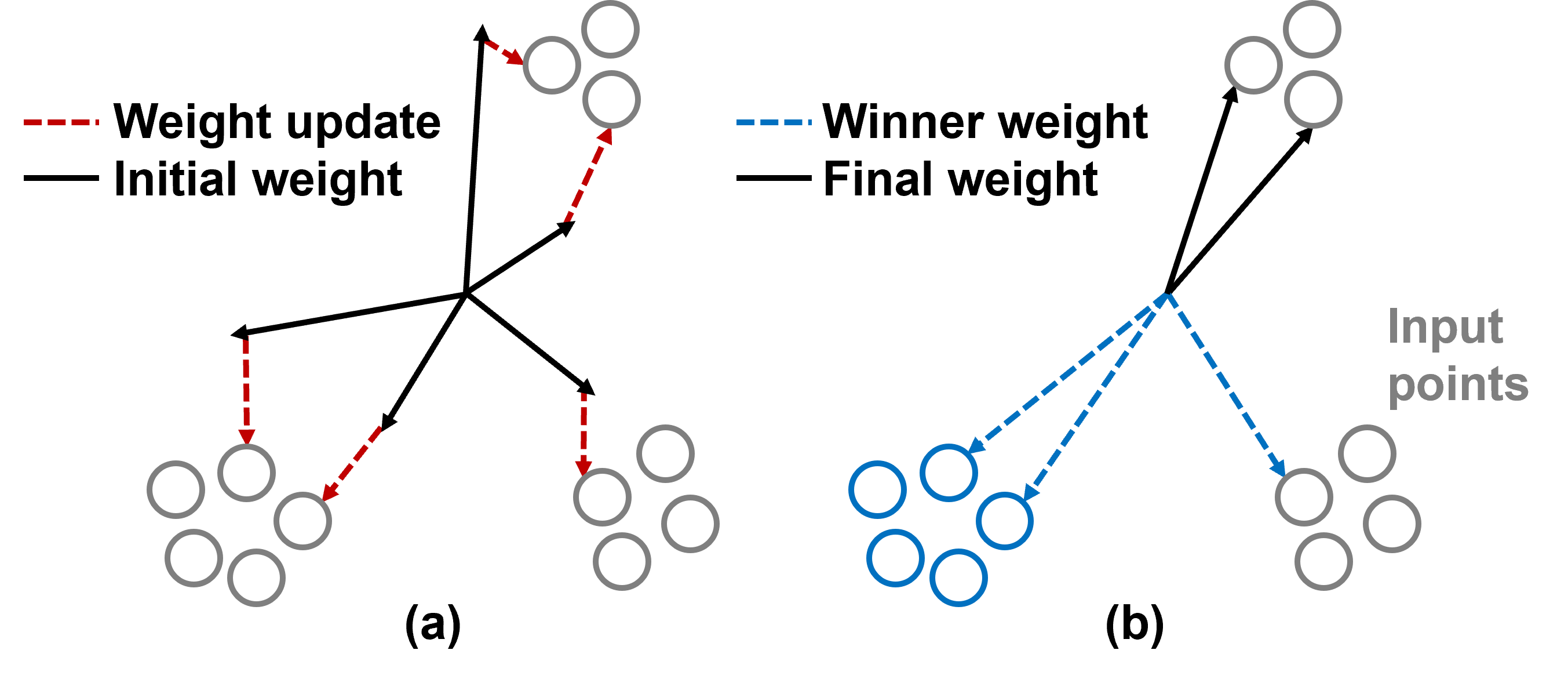}}
\caption{(a) describes the Hebbian learning rule in equation (\ref{equation_learning_rule}). (b) shows that the final weights converge to the nearest input points, and shows the weight vectors of winner neurons by the distance based k-WTA.}
\label{figure_Hebbian_WTA}
\vspace{-10pt}
\end{figure}

Fig. \ref{figure_Hebbian_WTA}(a) describes our Hebbian learning rule. The direction of weight updates indicates the closest input points from each weight vectors. The final location of weight vectors will converge to the closest points as described in Fig. \ref{figure_Hebbian_WTA}(b). 

\subsection{k-Winner-Takes-All on Point Sets}
k-WTA introduces a non-parametric way to form the sparse representation of latent vectors by inhibiting the neurons other than winners. Conventional WTA algorithm selects the winner neuron based on the highest output value. The winner is determined by \(\textbf{x} \cdot \textbf{w}=|\textbf{x}||\textbf{w}|cos\theta\) where \(\textbf{x}\) is a positional vector of a point. However, if the cosine similarity of points is not clearly different, the same neuron with the largest \(|\textbf{w}|\) will repeatedly win. For example in StarCraft II, the territories of two players can be assigned at the left bottom corner (origin) and right top corner, respectively. In this case, the largest weight vector pointing the right top corner from the origin will be always the winner. From the problem, we define the winner as the neuron with the lowest Euclidean distance \(|\textbf{x}-\textbf{w}|\). Then, the territories in the left bottom and right top corner are easily differentiable. We assume k=1 as only the lowest Euclidean distance is considered in this case. However, more weight vectors can be considered by changing the k value. Thus, we can control how many neurons to be used for the spatial attributes of points.

As Hebbian learning moves the weight vectors to be spatially close with points, the distance based k-WTA is a natural way to differentiate the points with different neurons. We adapt the k-WTA algorithm after the ReLU activation in the encoder, so the positional vector of each point is transformed to the latent vector with k non-zero elements. For example in Fig \ref{figure_Hebbian_WTA}(b), there are 5 weight vectors, so 5 neurons can have non-zero values without k-WTA. With k-WTA, blue points use the three closest weight vectors (three neurons) if we set k=3 regardless of the total number of neurons.

\subsection{Hebbian and anti-Hebbian Learning}
\begin{figure}
\centerline{\includegraphics[width=\columnwidth]{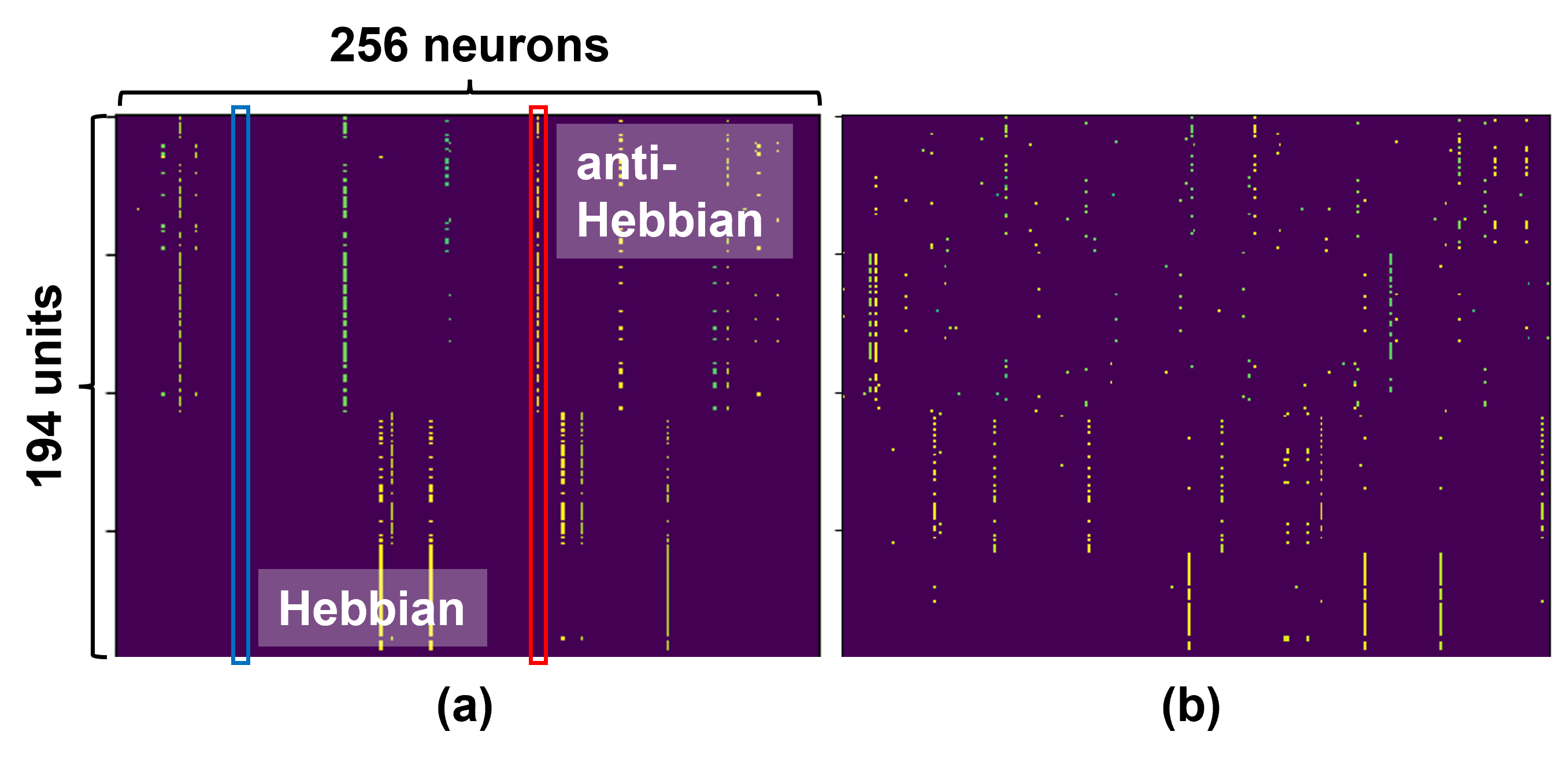}}
\caption{(a) is the visualization of latent vectors from each unit before training. We define two types of learning based on the neuron activity. (b) shows the latent vectors after the Hebbian learning. Activated neurons are more distributed compared to (a).}
\label{figure_Latent_Vector}
\vspace{-10pt}
\end{figure}

Fig. \ref{figure_Latent_Vector}(a) visualizes the output from the last MLP layer in the encoder before the Hebbian learning. The input is a (194, 2) matrix which is the x and y position of 194 units in a map, and the output before MaxPooling layer is a (194, 256) matrix. Dark background of the matrix indicates zero values, and each row has k=3 non-zero values where the colors are bright yellow or green. We observe that some units are clearly differentiated by different neurons with randomly initialized weights. However, multiple rows have same neuron activities, hence it is hard to distinguish them. It implies some weights and points are initially close each other. If we use Hebbian learning on it, the weight vectors will be closer to the points, and these points will activate same few neurons at the end. This will introduce the loss of local features as the next MaxPooling layer takes the single maximum value from these few neurons. From the observation, we assume the loss of location features can be reduced if more number of different neurons are remained after MaxPooling layer. Thus, the neurons that never activate should be closer to points while the weight vectors of too frequently activated neurons should be further away from points. We further develop our Hebbian learning rule with the consideration of the neuron activity. For example, we adapt Hebbian learning for the neuron in the blue box in Fig. \ref{figure_Latent_Vector}(a) as it never activates. However, anti-Hebbian learning is used for the neuron in the red box to move the weight vector far away from points.

\begin{equation}
\Delta\textbf{w}_{j} = \eta g(p_{j})\frac{1}{N}\sum_{i}{f(|\textbf{x}_{i}-\textbf{w}_{j}|)(\textbf{x}_{i}-\textbf{w}_{j})}
\label{equation_learning_rule2}
\end{equation}

\begin{equation}
  g(p_{j}) =  
    \begin{cases}
      1 & \text{if} \: p_{j} < p^{*} \\
      -1 & \text{if} \: p_{j} > p^{*} \\
      0 & \text{otherwise}
    \end{cases}       
\label{equation_Hebbian_antiHebbian}
\end{equation}

\begin{equation}
p_{j} = \frac{1}{N}\sum_{i}{u(y_{ij})} 
\label{equation_neuron_activity}
\end{equation}

We design a neuron activity aware Hebbian learning rule in equation (\ref{equation_learning_rule2}). It has an additional function \(g(p_{j})\), and it enables Hebbian and anti-Hebbian learning based on the different sign of output. The positive sign of \(g(p_{j})\) makes the learning rule to be same as equation (\ref{equation_learning_rule}) in the previous section. If the sign of \(g(p_{j})\) is negative, we define it as anti-Hebbian learning indicating the weight vectors move further away from points. \(p_{j}\) is the activity of \(j^{th}\) neuron defined by the fraction of the number of neuron activation and the total number of points. In equation (\ref{equation_neuron_activity}), \(y_{ij}\) is the output of \(i^{th}\) point and \(j^{th}\) neuron, and \(u\) is a step function as we are interested in whether the neuron is activated or not. We assume that there is desired neuron activity \(p^{*}\) to reduce the information loss, hence we determine the type of learning by comparing \(p_{j}\) and \(p^{*}\) in equation (\ref{equation_Hebbian_antiHebbian}).

\begin{figure*}
\includegraphics[width=\textwidth]{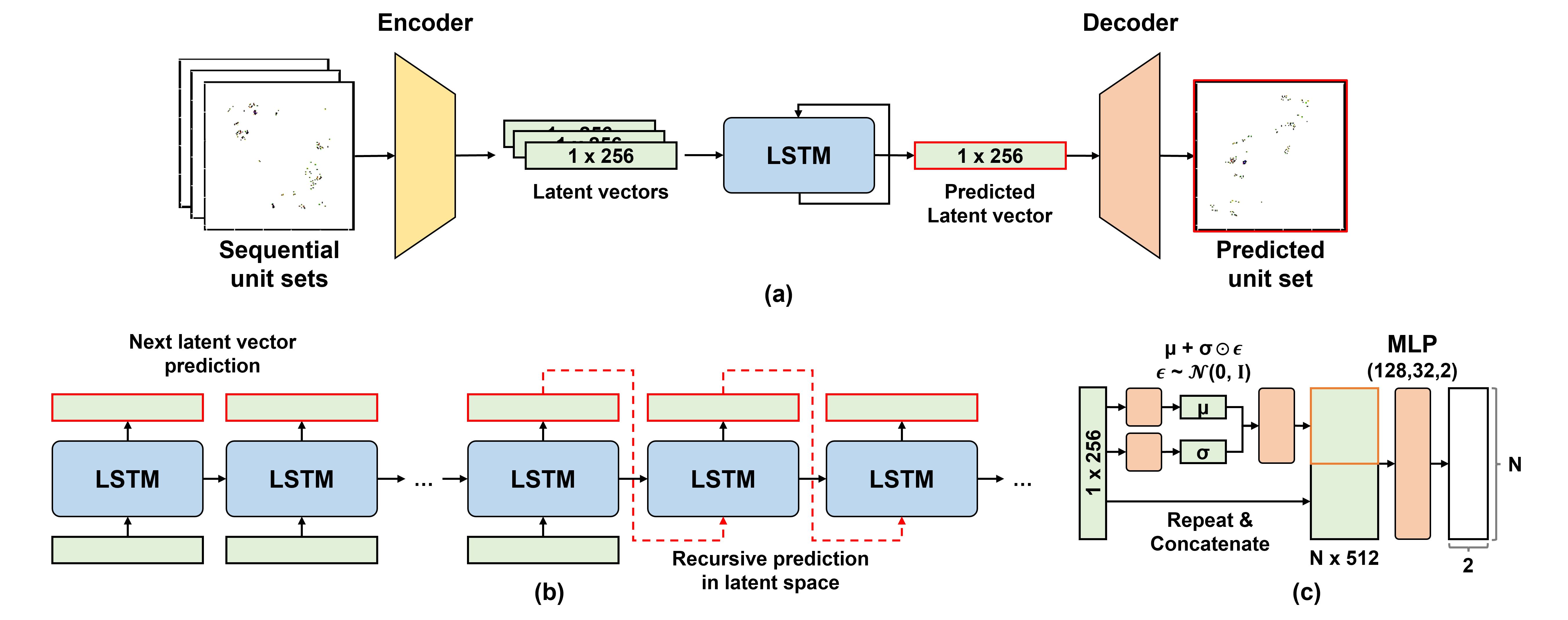}
\caption{(a) shows the schematic of overall network architecture including encoder, LSTM, and decoder. (b) describes the data flow in the LSTM. There are two types of prediction, direct prediction and recursive prediction. (c) shows the detailed decoder design to sample the variable number of unit position from the latent vector.}
\label{figure_Network_Architecture}
\vspace{-10pt}
\end{figure*}

\subsection{Optimal Neuron Activity}
We derive the optimal neuron activity \(p^{*}\) based on the intuition that the loss of local features can be reduced when the latent vectors are orthogonal. In the case, more number of different neurons are remained to represent the local features after MaxPooling layer. Thus, our objective function follows:
\begin{equation}
\mathrm{min} \sum_{i, j, i \neq j}{\textbf{x}_{i} \cdot \textbf{x}_{j}}
\label{equation_inner_product}
\end{equation} 
where \(\textbf{x}_{i}\) and \(\textbf{x}_{j}\) are the latent vectors of \(i^{th}\) and \(j^{th}\) points. As we only consider the activation of neurons, not the output values, each element of the latent vectors are assumed to be binary. Then, the sum of inner products can be considered as counting the pairs of ones with same index in all latent vectors. We can rewrite the equation (\ref{equation_inner_product}) as:
\begin{equation}
\mathrm{min} \sum_{r}{_{n_{r}}C_{2}}=\mathrm{min} \sum_{r}{\frac{n_{r}^{2}-n_{r}}{2}}
\label{equation_combination}
\end{equation} 
where \(n_{r}\) is the number of the \(r^{th}\) neuron activation, so it ranges from 0 to the number of points (\(N\)). We apply k-WTA on each latent vector, so the total number of neuron activations is \(k\) times \(N\). Hence, a boundary condition is obtained as:
\begin{equation}
\sum_{r}{n_{r}}=Nk
\label{equation_boundary_condition}
\end{equation} 
From the boundary condition and Lagrange multiplier, we define the following equation.
\begin{equation}
\sum_{r}{\frac{n_{r}^{2}-n_{r}}{2}}-\lambda(\sum_{r}{n_{r}}-Nk)=0 
\label{equation_lagrange}
\end{equation} 
Equation (\ref{equation_lagrange}) is satisfied when the solution \(n_{r}^{*}\) is \(\frac{Nk}{d}\) where d is the number of neurons. Thus, the optimal neuron activity \(p^{*}\) is defined by:
\begin{equation}
p^{*}=\frac{n_{r}^{*}}{N}=\frac{k}{d}
\label{equation_optimal_activity}
\end{equation} 

Fig. \ref{figure_Latent_Vector}(b) shows the latent vectors after the neuron activity aware Hebbian learning. Compared to Fig. \ref{figure_Latent_Vector}(a), we observe more number of neurons are used to represent the latent vectors. In particular, the activity of the neuron in the blue box increases, and no neurons activate as frequent as the neuron in the red box after the training.

\section{Overview}
\subsection{Dataset}
We extract our dataset from open source replay videos provided by Blizzard. Replay packs can be found on GitHub (https://github.com/Blizzard/s2client-proto\#downloads). There are the three types of races in the game, but we only use a Terran versus Terran replay game. We randomly select a game from the replay pack1 in the source. The position of all units is extracted every \(0.2\) seconds, and units from two players are incorporated together in each data. Other information such as buildings and resources are not considered. We extract the spatial information of units as two types. One is set based x and y position which is (N, 2) dimension where \(N\) is the number of units in a map. As we set the resolution of a map as (256, 256), x and y position values are divided by 255 to normalize from 0 to 1. Another type is (256, 256) image data which pixel values are binary. Most of pixels are zero, but ones if there are units on the pixels. We use PySC2 package to extract the data from replay videos. PySC2 is a Python based StarCraft II learning environment developed by DeepMind \cite{PySC2}. Entire extracted data includes 11,000 frames. We transform the sequential frames as 220 chunks for LSTM training. Each chunks has successive 50 frames with 10 seconds duration. Training data is defined by 80\% of the randomly shuffled chunks, and the other two 10\% of the chunks are used for validation and test data, respectively.

\subsection{Model Architecture}
Fig. \ref{figure_Network_Architecture}(a) shows the overview of entire model architecture to predict the future location of units. There are total three modules including encoder, LSTM, and decoder. The encoder is explained in the previous section with Fig. \ref{figure_Encoder_Architecture}. Our decoder is motivated from variational autoencoder, and it decodes a latent vector to a unit set. Fig. \ref{figure_Network_Architecture}(c) describes the schematic of the decoder module. We design a sampling layer to generate a variable number of elements in the set. Four light orange blocks in Fig. \ref{figure_Network_Architecture}(c) indicate MLP layers, and ReLU is used as an activation function. The model learns the dynamics of a point set using LSTM in latent space. Fig. \ref{figure_Network_Architecture}(b) shows the data flow in our single-layer and unidirectional LSTM module. It consists of 64 hidden and cell states, and each of fully-connected layers is attached before and after the LSTM cell to match the dimension with latent vectors.

\subsection{Training Procedure}
We first train the encoder with the Hebbian learning and k-WTA. Other modules are not required since we do not use supervised or self-supervised learning methods for the encoder. After the encoder training, the decoder is trained with the pre-trained encoder. Here, the encoder is frozen, hence only parameters of the decoder are updated. At this point, the LSTM is not yet combined with the encoder and decoder. The decoder learns to reconstruct sets from latent vectors by self-supervised learning. By training the encoder and decoder without the LSTM, training data does not require to consider the sequences of sets. We can randomly shuffle the training data instead of using chunks during learning the reconstruction. 

Next, we insert the LSTM between the pre-trained encoder and decoder to learn the dynamics of a point set in latent space. Fig. \ref{figure_Network_Architecture}(b) describes the data flow in the LSTM. Given the chunk with successive 50 sets, the LSTM observes the first 25 latent vectors of corresponding sets, and predicts the latent vector of a next time step every observation. After the observation, the LSTM repeatedly uses the last prediction as an input to predict the next 24 time steps. Thus, the LSTM and decoder generates 49 sets as output, and we calculate the loss from each of the sets. Similar to the decoder training, the parameters of the encoder and decoder are frozen.  

While the encoder does not require the loss function, the decoder and LSTM are trained by the error from the output of the decoder. We use the same loss function defined in equation (\ref{equation_loss}) for the two modules. The output generates the point set where each of elements are two-dimensional position of units, and the corresponding ground truth is also a positional set. Our loss function is based on Chamfer loss which calculates the distance between the two sets in a permutation invariant way.
\begin{equation}
\mathcal{L}{(S_{1}, S_{2})}=\sum_{x \in S_{1}}{\underset{y \in S_{2}}{\mathrm{min}} \: d(x,y)} + \sum_{y \in S_{2}}{\underset{x \in S_{1}}{\mathrm{min}} \: d(x,y)}
\label{equation_loss}
\end{equation} 
\begin{equation}
d(x, y)=
    \begin{cases}
      0.5(x-y)^{2} & \text{if} \: |x-y| < 1 \\
      |x-y|-0.5 & \text{otherwise}\\
    \end{cases}
\label{equation_smoothl1}
\end{equation} 
Chamfer loss is generally based on squared Euclidean distance. Given the units are moving in a large map, we instead use the smooth L1 distance \(d(x, y)\) in equation (\ref{equation_loss}) so that the gradient is less sensitive to outliers.

\begin{figure}
\centerline{\includegraphics[width=\columnwidth]{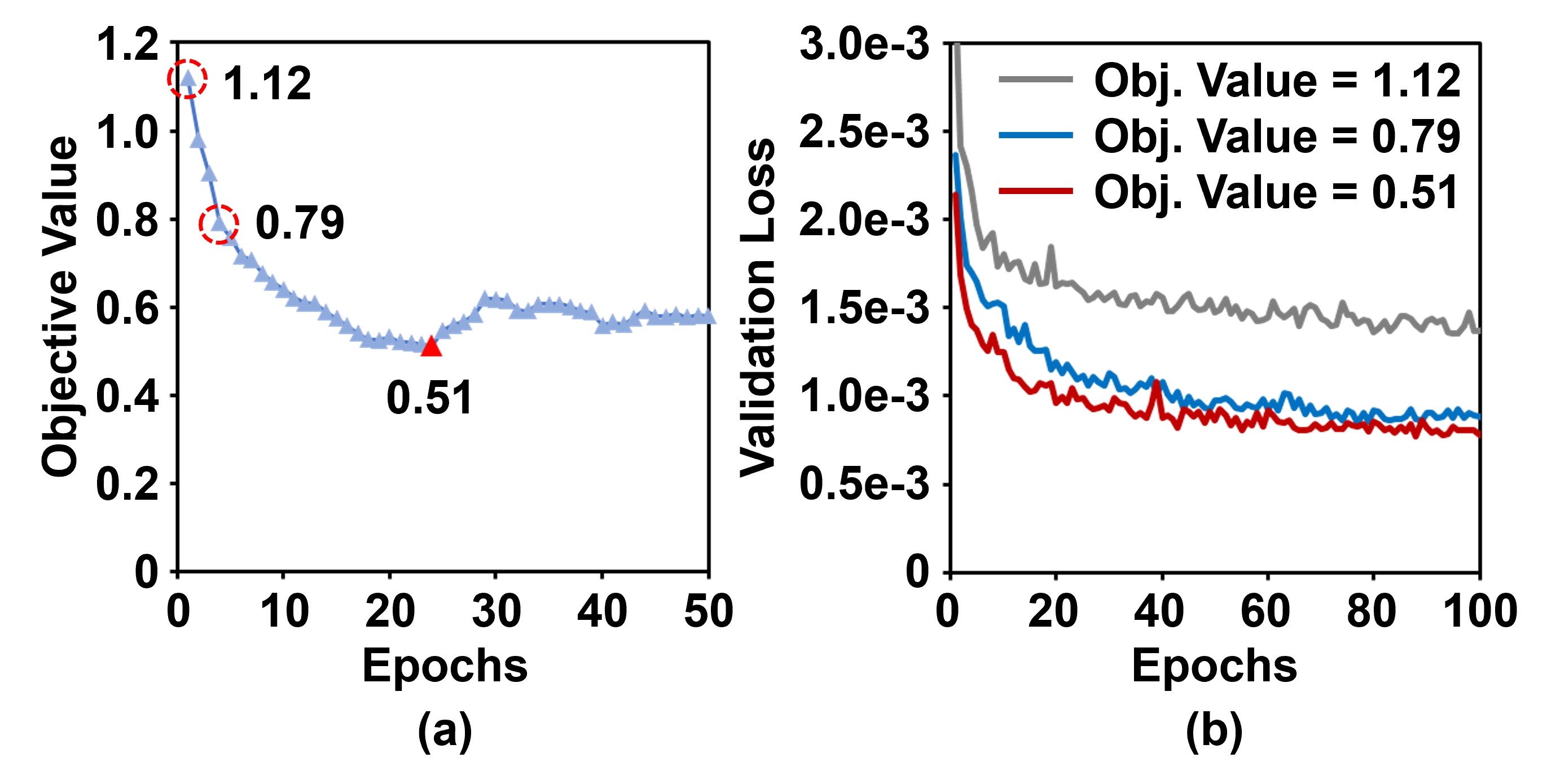}}
\caption{(a) shows the change of the objective value in equation (\ref{equation_combination}) by the Hebbian learning with 1-WTA. The minimum value is marked by the red triangle. (b) compares the reconstruction loss of the three models with different objective values.}
\label{figure_Objective}
\vspace{-10pt}
\end{figure}

\begin{table}
\caption {Reconstruction Loss with Various Models} \label{tab:title}
\centering
\begin{tabular}{>{\hspace{0pt}}m{0.24\linewidth}>{\hspace{0pt}}m{0.15\linewidth}>{\hspace{0pt}}m{0.15\linewidth}>{\hspace{0pt}}m{0.15\linewidth}} 
\hline
\textbf{Model} & \textbf{Objective Value} & \textbf{Train Loss} & \textbf{Validation Loss} \\ 
\hline
Untrained & - & 1.073e-3 & 1.049e-3 \\
Self-supervised & - & 5.738e-4 & 6.261e-4 \\
\hline
1-WTA & 1.122 & 1.213e-3 & 1.351e-3 \\
\textbf{1-WTA+Hebbian} & \textbf{0.513} & \textbf{6.418e-4} & \textbf{7.677e-4} \\
\hline
3-WTA & 2.391 & 8.207e-4 & 9.009e-4 \\
\textbf{3-WTA+Hebbian} & \textbf{1.126} & \textbf{5.846e-4} & \textbf{7.110e-4} \\
\hline
5-WTA & 3.271 & 7.626e-4 & 8.633e-4 \\
\textbf{5-WTA+Hebbian} & \textbf{1.890} & \textbf{5.517e-4} & \textbf{6.855e-4} \\
\hline
7-WTA & 8.938 & 6.873e-4 & 7.641e-4 \\
\textbf{7-WTA+Hebbian} & \textbf{2.517} & \textbf{6.213e-4} & \textbf{7.314e-4} \\
\hline
\end{tabular}
\label{table_reconstruction_loss}
\vspace{-10pt}
\end{table}

\section{Experimental Results}
In this section, we first verify the effectiveness of the Hebbian learning in terms of the set reconstruction and prediction, and then check how the Hebbian learning performs with limited training data for the encoder compared to self-supervised learning. Lastly, we compare our model against a frame-based approach with focus on the computational complexity. For baseline models, untrained and self-supervised encoder is used with the same architecture. There is another related work on the spatiotemporal point set in StarCraft II \cite{TemporalPoint}. However, it is limited to predict the movement of the fixed number of units. For the reason, ConvLSTM \cite{ConvLSTM} is used as a frame-based approach to compare with our entire model. 

\begin{figure*}
\includegraphics[width=\textwidth]{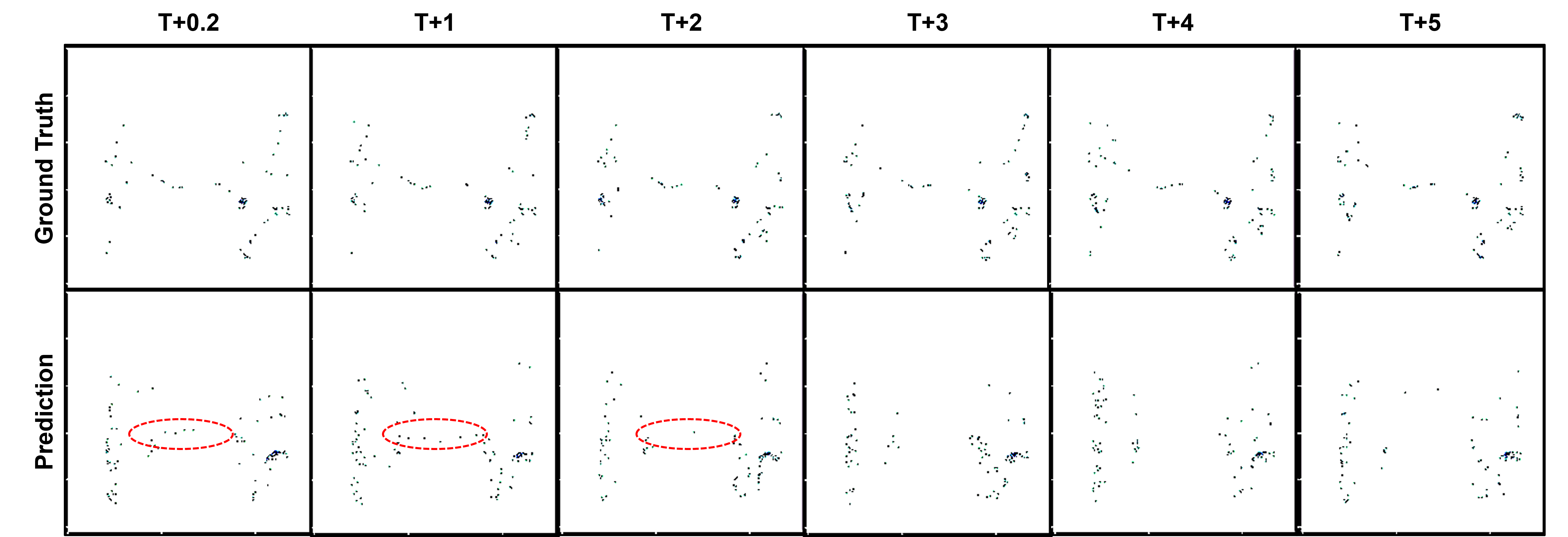}
\caption{Predicted frames and corresponding ground truth from the test dataset. First prediction at T+0.2 is based on the latent vector from the encoder, and later prediction uses the output from the LSTM. In all figures, the values of pixels are binary where bright pixels are zeros.}
\label{figure_Recursive_Timestep}
\vspace{-10pt}
\end{figure*}

\subsection{Effectiveness of Hebbian Learning: Set Reconstruction}
\begin{figure}
\centerline{\includegraphics[width=\columnwidth]{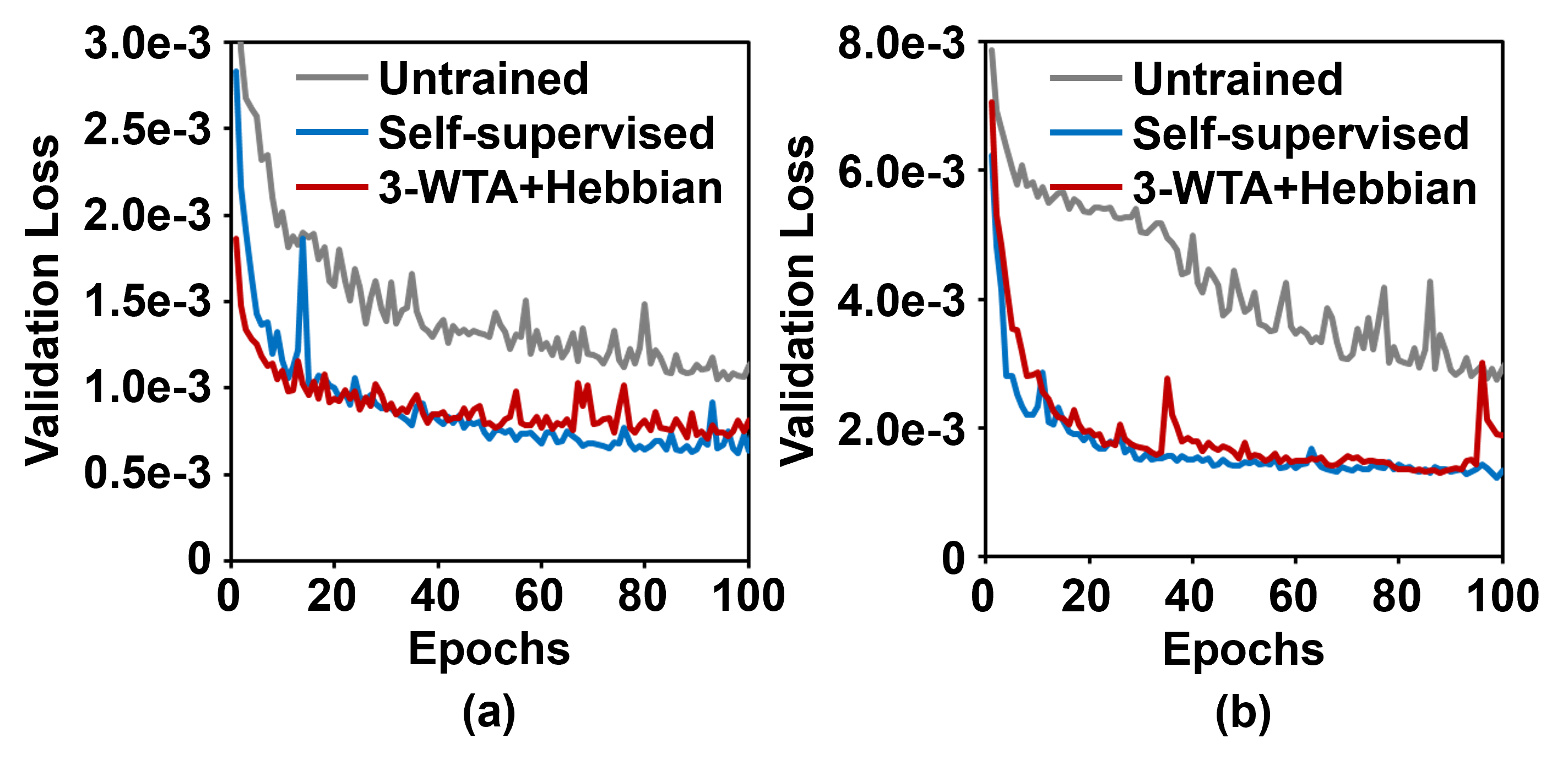}}
\caption{(a) compares the reconstruction loss with untrained, self-supervised, and unsupervised models. (b) compares the prediction loss with the same models in (a).}
\label{figure_Learning_Speed}
\vspace{-10pt}
\end{figure}

As we first train the encoder and decoder, we compare the reconstruction loss to verify the effectiveness of the neuron activity aware Hebbian learning. Here, the reconstruction means decoding the latent vector to the input point set. Before then, we investigate whether the Hebbian learning minimizes the objective function in equation (\ref{equation_combination}), and whether minimizing the objective function is valid approach to improve the encoding performance. We simply calculate the average objective value without the second term in equation (\ref{equation_combination}) since it is constant during the training due to the boundary condition in equation (\ref{equation_boundary_condition}). Also, the objective function of the last layer in the encoder is considered. Fig. \ref{figure_Objective}(a) shows the changes of the objective value during the 50 epochs of Hebbian learning. We set the learning rate 0.01, and training batch 16. We observe that the Hebbian learning reduces the objective value, but sometimes increases the value. As the learning rule in equation (\ref{equation_learning_rule2}) does not directly optimize the objective value, we pick the model with the minimum value which is the red point in the Fig. \ref{figure_Objective}(a). We observe the Hebbian learning generally arrives at the minimum within 50 epochs in k=1, 3, 5, and 7 settings. 

We save the three models with the different objective values during the 50 epochs to compare the reconstruction loss of them. Fig. \ref{figure_Objective}(b) compares the reconstruction loss of the three models with the objective value 1.12, 0.79, and 0.51. The values of corresponding data points are written in Fig. \ref{figure_Objective}(a). The encoder is pre-trained by the Hebbian learning, and then the parameters of the encoder are frozen. Hence, the epochs in Fig. \ref{figure_Objective}(b) are the training epochs of the decoder. We set learning rate 0.0005 with Adam optimizer, training batch 16, and test batch 256. We observe that there is no linear relation between the loss and the objective value, but the validation loss is lower in the higher objective value.

In addition, we compare the reconstruction loss with different k values to further verify the effectiveness of the objective function and the Hebbian learning. Table \ref{table_reconstruction_loss} summarizes the reconstruction loss of various models, and the minimum values during 100 training epochs are listed. Untrained and self-supervised models are also incorporated as baseline models. In the untrained model, the decoder is trained, but the encoder is untrained with randomly initialized weights. We train the self-supervised model by end-to-end training through a backpropagation algorithm. Here, the self-supervised model can be also regarded as unsupervised, but we differentiate it to avoid the confusion. Both models have the same model architecture with the previous experiments. The only difference is that k-WTA is not adapted. In all different k values, the Hebbian learning reduces the initial objective values, and the validation loss also decreases. The validation loss is the lowest with k=5 in unsupervised models. We observe that self-supervised model has lower validation loss compared to all unsupervised models, but the difference of the loss is within few percentages.
\begin{table}
\caption {Prediction Loss with Various Models} \label{tab:title}
\centering
\begin{tabular}{>{\hspace{0pt}}m{0.25\linewidth}>{\hspace{0pt}}m{0.15\linewidth}>{\hspace{0pt}}m{0.15\linewidth}>{\hspace{0pt}}m{0.15\linewidth}}
\hline
\textbf{Model} & \textbf{Train Loss} & \textbf{Validation Loss} & \textbf{Test Loss}  \\ 
\hline
Untrained & 2.736e-3 & 2.764e-3 & 3.114e-3 \\
Self-supervised & 1.266e-3 & 1.182e-3 & 1.322e-3 \\
\hline
1-WTA+Hebbian   & 1.193e-3 & 1.174e-3 & 1.222e-3 \\
3-WTA+Hebbian & 1.168e-3 & 1.118e-3 & 1.309e-3 \\
\textbf{5-WTA+Hebbian} & \textbf{1.159e-3} & \textbf{1.107e-3} & \textbf{1.156e-3} \\
7-WTA+Hebbian & 1.448e-3 & 1.425e-3 & 1.574e-3 \\
\hline
\end{tabular}
\label{table_prediction_loss}
\vspace{-10pt}
\end{table}

\subsection{Effectiveness of Hebbian Learning: Set Prediction}
From the data flow of the LSTM in Fig. {\ref{figure_Network_Architecture}}(b), we predict the latent vector of the next time step until the half of frames is observed, and then recursively predict the latent vectors using the previously predicted vector. At this time, We assume that the number of units in a map is known to simplify the decoding process. We first verify whether the Hebbian learning loses the temporal information underlying the sequential point set as we only compare the reconstruction performance in the previous section. Fig. \ref{figure_Learning_Speed}(b) compares the prediction loss with untrained, supervised, and the Hebbian learning models. The reconstruction loss of the same models are shown in Fig. \ref{figure_Learning_Speed}(a) to compare the change of loss resulted by inserting the LSTM between the encoder and decoder. Here, the encoder and decoder are pre-trained without the LSTM, and their parameters are not updated during the LSTM training. Both the self-supervised and Hebbian learning models show the validation loss is quickly saturated, and increased compared to the reconstruction loss. The prediction loss of more various models are summarized in Table \ref{table_prediction_loss}. We observe that the prediction loss in the Hebbian learning models with k=1, 3, and 5 is less increased compared to the baselines. Furthermore, the k=5 case shows the lower test loss than the self-supervised model.

Fig. \ref{figure_Recursive_Timestep} shows the repeatedly predicted sets. The ground truth and prediction have binary values where dark pixels are ones indicating the presence of units. The output from our model is originally a positional set, and we plot the corresponding map in the figure. As described in Fig. \ref{figure_Network_Architecture}(b), the LSTM recursively predicts the latent vectors by using the output in the previous time step. The LSTM observes 25 frames in the past, and predicts the rest 25 frames. Hence, the figure shows the 1st, 5th, 10th, 15th, 20th, and 25th predicted frames from left to right. We observe that the movement of some units is appropriately predicted in the early time steps. There are few units near the center of a map in the ground truth, and they are moving from the left to the right terrain. Also, in the red circle at T+0.2, the predicted units match with the ground truth. At T+1 and T+2, the units are predicted at the similar location and we can see the moving route of them. However, the failure of the prediction is observed after then. It is generally known in LSTM that the repeated prediction accumulates the error with the loss of features. We plot how the loss varies during the entire prediction in the next section.

\begin{figure}
\centerline{\includegraphics[width=\columnwidth]{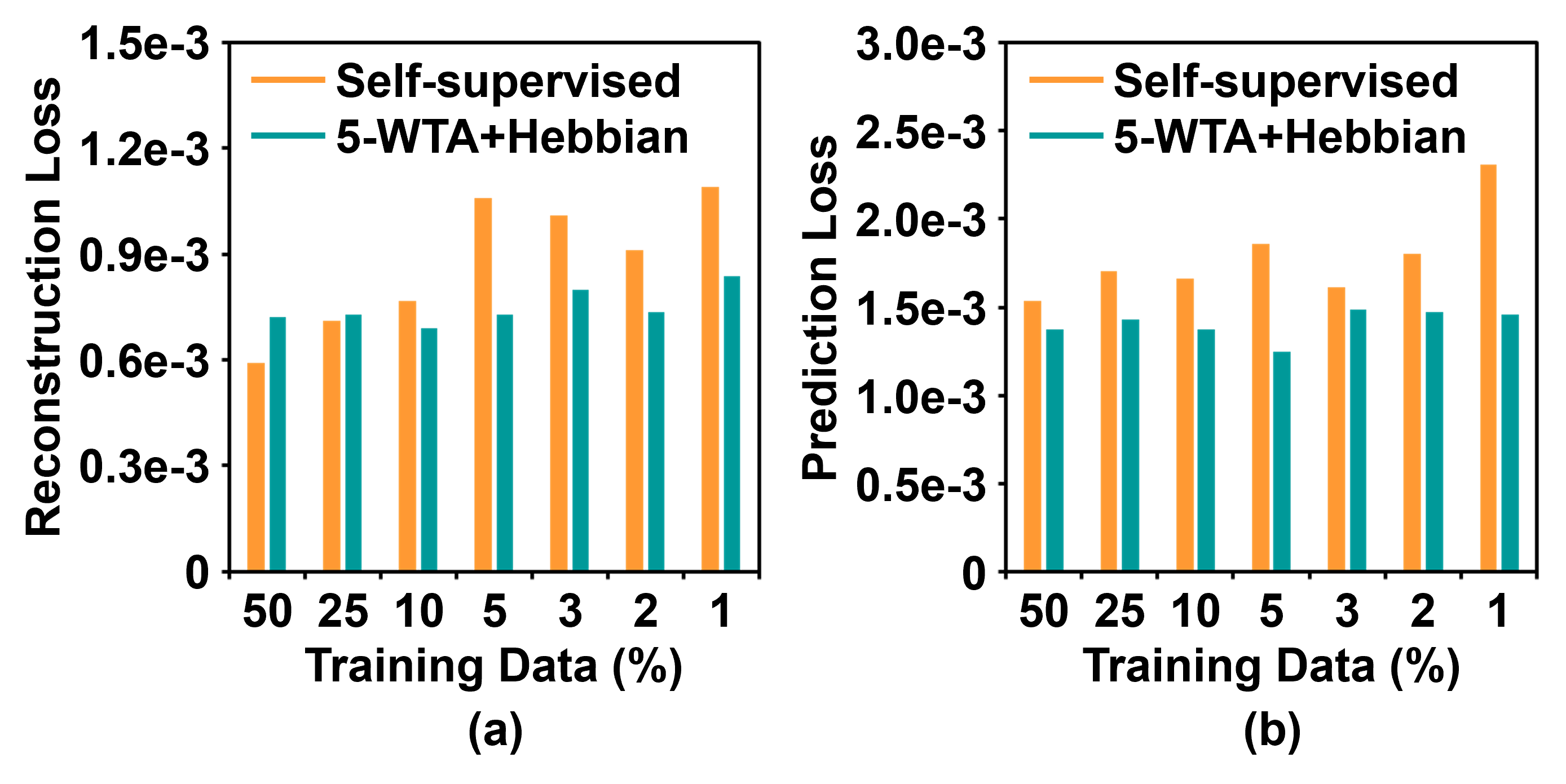}}
\caption{(a) shows the change of reconstruction loss in self-supervised and unsupervised models. (b) shows the change of prediction loss in the same models. Both plots are based on the test dataset}
\label{figure_Limited_Data}
\vspace{-10pt}
\end{figure}

\begin{table}
\caption {Comparison against ConvLSTM} \label{tab:title}
\centering
\begin{tabular}{>{\hspace{0pt}}m{0.25\linewidth}>{\hspace{0pt}}m{0.15\linewidth}>{\hspace{0pt}}m{0.15\linewidth}>{\hspace{0pt}}m{0.15\linewidth}}
\hline
\textbf{Model} & \textbf{Parameters} & \textbf{Activations} & \textbf{FLOPs} \\
\hline
Encoder & 36.9k & 132.5k & 12.1M \\
LSTM & 66.4k & 0.8k & 131.1k \\
Decoder & 71.7k & 173.6k & 22.2M \\
\hline
\textbf{5-WTA+Hebbian} & \textbf{175.0k} & \textbf{307.0k} & \textbf{34.4M} \\
ConvLSTM & 208.7k & 33.7M & 27.3G \\
\hline
\end{tabular}
\label{table_ConvLSTM}
\vspace{-10pt}
\end{table}

\subsection{Limited Training Data for Encoder}
Recent studies on Hebbian learning \cite{SampleEfficientHebbian} and its variant Spike Timing Dependent Plasticity (STDP) learning \cite{HSNN} have shown the feasibility of data efficient training. We also investigate the data efficiency of our Hebbian learning rule compared to self-supervised learning. Fig. \ref{figure_Limited_Data} shows the test loss for the set reconstruction and prediction with the limited amount of training data. Here, the limited data is only for the encoder as other modules are trained by self-supervised learning. The LSTM and decoder are trained with full data. We observe that self-supervised learning introduces the large degradation of the loss in both reconstruction and prediction. In particular, the reconstruction loss in the self-supervised model starts to be similar to the one in the untrained encoder if the amount of training data is less than \(10\%\). However, the unsupervised learning has no remarkable degradation except the \(1\%\) data case. The unsupervised learning still performs better in the prediction task with the limited data while the considerable increase of the loss is observed in self-supervised learning. Thus, our Hebbian learning rule also has the benefits of the data efficient training with the less performance degradation.

\subsection{Comparison with Frame-based approach}
\begin{figure}
\centerline{\includegraphics[width=\columnwidth]{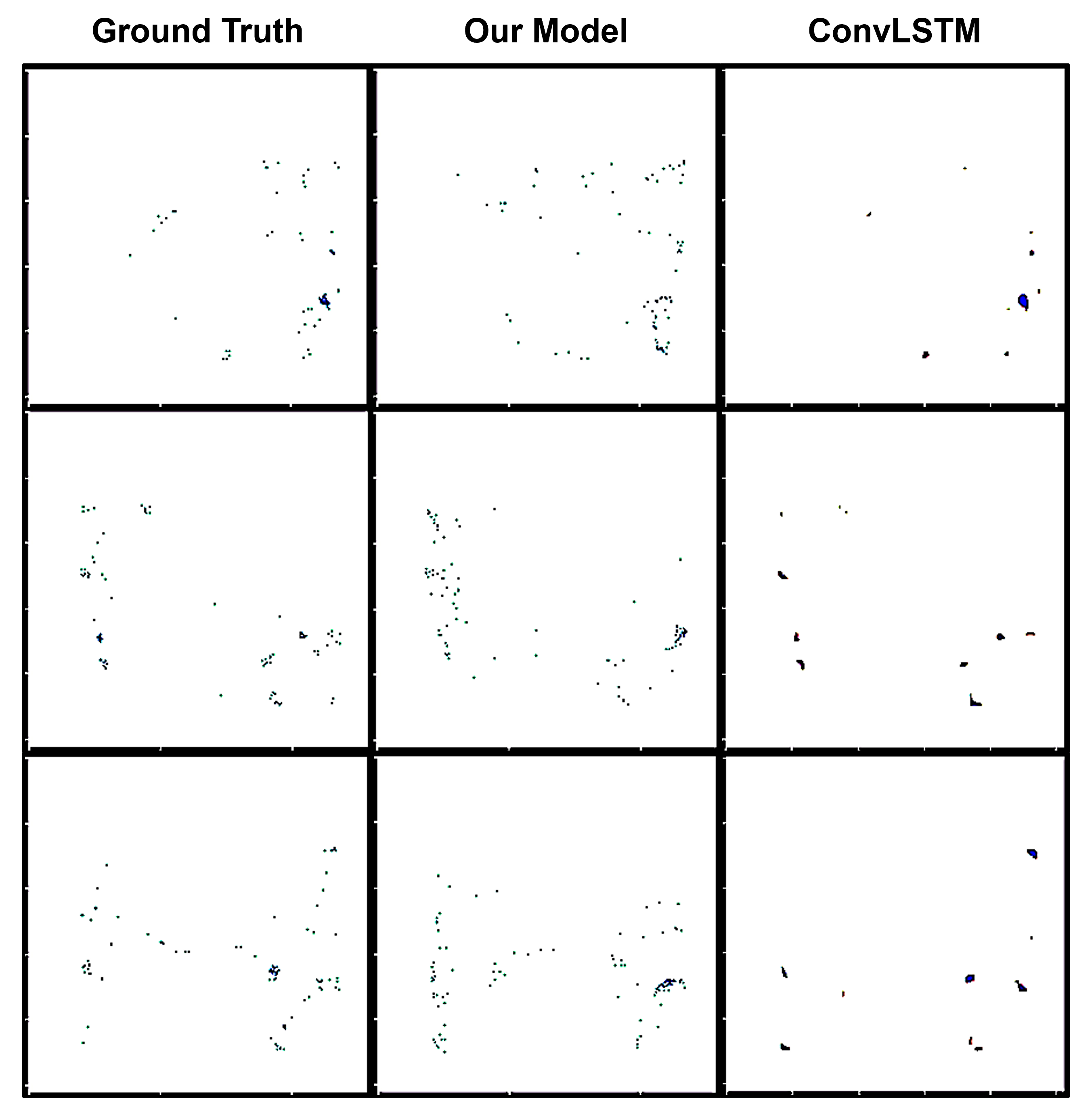}}
\caption{Three examples of predicted 26th frames after observing past 25 frames. All the figures are based on the test dataset. First column shows the ground truth of three different examples. Next columns are the predicted frames from our 5-WTA+Hebbian model and the ConvLSTM.}
\label{figure_Next_Timestep}
\vspace{-10pt}
\end{figure}

We design a single-layer and unidirectional ConvLSTM model to have the similar number of trainable parameters with our model. The input and hidden state have \((1, 256, 256)\) and \((32, 256, 256)\) dimension, respectively. The output from the ConvLSTM is followed by an additional convolutional layer to generate a \((1, 256, 256)\) map same as the input dimension. Each of four gates in the ConvLSTM has a \((7, 7, 33, 32)\) kernel, and the last convolutional layer has a \((7, 7, 32, 1)\) kernel. The ConvLSTM is trained with the same data flow, and binary cross entropy is used as a loss function to predict the presence of units by pixel-wise probabilities. 

Table \ref{table_ConvLSTM} compares the computational complexity of our model and the ConvLSTM. We approximately estimate FLOPs per prediction with focus on fully-connected layers and convolution layers as they consume most of the computation. Also, we consider the total summation of output dimensions in these layers and activation functions to calculate the activations. We observe that our model has \(\times110\) less activations and \(\times793\) less FLOPs compared to the ConvLSTM with the similar number of trainable parameters. Our model obtains the considerable computational benefits mainly due to the dimensional reduction by the encoder. It is important to note that the LSTM in our model learns the dynamics of latent vectors with \((1, 64)\) dimension while the ConvLSTM performs with \((32, 256, 256)\) tensors. As the FLOPs are based on single prediction, the difference of the computations in the two models will increase during the repeated prediction.

\begin{figure}
\centerline{\includegraphics[width=\columnwidth]{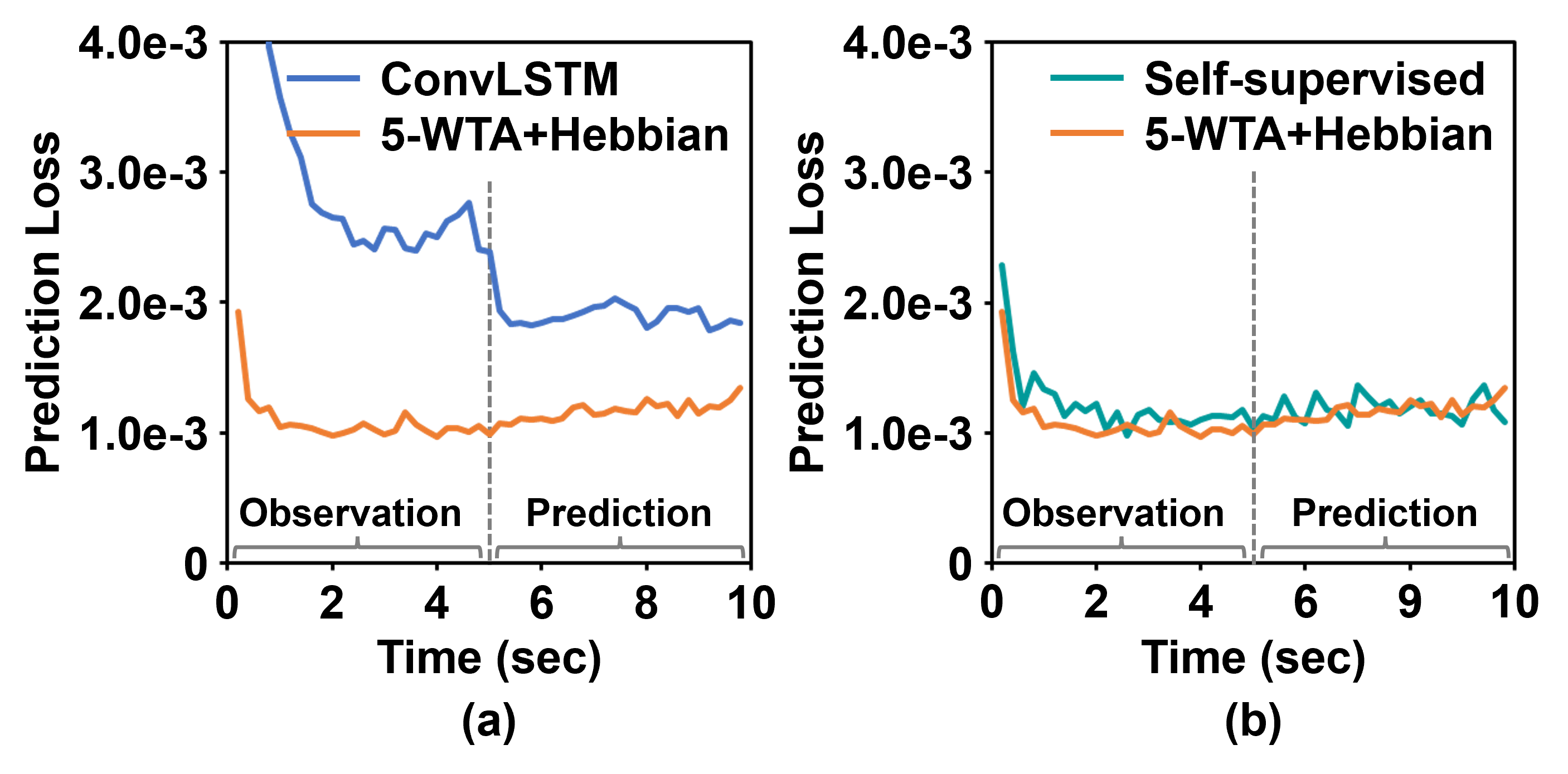}}
\caption{(a) compares the test loss of our 5-WTA+Hebbian model and the ConvLSTM. Total duration is 10 seconds, and the two models repeatedly predict the frames after 5 seconds. Both models show large losses in early time steps due to the absence and lack of the observation. (b) compares the same loss with the self-supervised model instead of the ConvLSTM.}
\label{figure_chamfer_mse_loss}
\vspace{-10pt}
\end{figure}

Fig. \ref{figure_Next_Timestep} shows the three examples of predicted frames from the two models and corresponding ground truth. The ConvLSTM predicts a probability map, and the map is filtered by selecting the same number of high-probability pixels as the number of units. In all cases, both models observe 25 frames, and then the predicted next frames are plotted. We observe that the prediction from the two models mainly differ in terms of outlier units. The ConvLSTM can predict the location of unit groups as shown in the figure, but other units that does not move as large groups are not predicted with high enough probabilities. Instead, our model can predict more number of the outliers though the predicted position is less accurate. The two models have pros and cons from the observation. However, from the perspective of the game, it is important to capture the interaction between two players to understand the game evolution, and it frequently incorporates the dynamics of the outlier units. For example in the third row of Fig. \ref{figure_Next_Timestep}, few units are on the middle area between the left and right group of units. It could be the situation that they move toward the terrain of another player to attack him. We see the predicted map of our model in Fig. \ref{figure_Next_Timestep} incorporates these units while the ConvLSTM completely fails to predict them.

Fig. \ref{figure_chamfer_mse_loss}(a) shows the comparison of Chamfer loss between the two models. The LSTM and ConvLSTM make the prediction during the observation as well, so the figure plots the loss values for 49 time steps. As expected from Fig. \ref{figure_Recursive_Timestep}, both models show the increasing loss with time during the repeated prediction. We observe that our model performs better in terms of the lower Chamfer loss. It might be interpreted along with the observation in Fig. \ref{figure_Next_Timestep}. Though the ConvLSTM makes the accurate prediction for the group of units than our model, the large distance errors from the outliers increase the entire distance errors in the ConvLSTM. Our model achieves average Chamfer loss of \(1.16\times10^{-3}\) for the test dataset. The ConvLSTM shows average Chamfer loss of \(3.25\times10^{-3}\). Fig. \ref{figure_chamfer_mse_loss}(b) shows the similar comparison between the self-supervised and unsupervised model, but no considerable difference is observed.

\section{Conclusion}
This paper presents the unsupervised learning method of a point set in StarCraft II units to efficiently and effectively learn the evolution of point sets. We discuss how the neuron activity and k-WTA can be combined with Hebbian learning for the point set representation. The prediction performance and computational cost are compared with self-supervised learning in the same model and ConvLSTM. Our Hebbian learning rule shows the lower distance errors with less activations and FLOPs than the baselines.

\bibliographystyle{IEEEtran}

\bibliography{Reference}

\begin{thebibliography}{10}
\providecommand{\url}[1]{#1}
\csname url@samestyle\endcsname
\providecommand{\newblock}{\relax}
\providecommand{\bibinfo}[2]{#2}
\providecommand{\BIBentrySTDinterwordspacing}{\spaceskip=0pt\relax}
\providecommand{\BIBentryALTinterwordstretchfactor}{4}
\providecommand{\BIBentryALTinterwordspacing}{\spaceskip=\fontdimen2\font plus
\BIBentryALTinterwordstretchfactor\fontdimen3\font minus
  \fontdimen4\font\relax}
\providecommand{\BIBforeignlanguage}[2]{{%
\expandafter\ifx\csname l@#1\endcsname\relax
\typeout{** WARNING: IEEEtran.bst: No hyphenation pattern has been}%
\typeout{** loaded for the language `#1'. Using the pattern for}%
\typeout{** the default language instead.}%
\else
\language=\csname l@#1\endcsname
\fi
#2}}
\providecommand{\BIBdecl}{\relax}
\BIBdecl

\bibitem{Atari}
V.~Mnih \emph{et~al.}, ``Playing atari with deep reinforcement learning,'' in
  \emph{arXiv preprint arXiv:1312.5602}, 2013.

\bibitem{Chess}
D.~Silver \emph{et~al.}, ``A general reinforcement learning algorithm that
  masters chess, shogi, and go through self-play,'' \emph{Science}, vol. 362,
  no. 6419, pp. 1140--1144, 2018.

\bibitem{Go}
------, ``Mastering the game of go with deep neural networks and tree search,''
  \emph{Nature}, vol. 529, no. 7587, pp. 484--489, 2016.

\bibitem{GrandMasterSC2}
O.~Vinyals \emph{et~al.}, ``Grandmaster level in starcraft ii using multi-agent
  reinforcement learning,'' \emph{Nature}, vol. 575, no. 7782, pp. 350--354,
  2019.

\bibitem{PySC2}
------, ``Starcraft ii: A new challenge for reinforcement learning,'' in
  \emph{arXiv preprint arXiv:1708.04782}, 2021.

\bibitem{DefogGan}
Y.~Jeong, H.~Choi, B.~Kim, and Y.~Gwon, ``Defoggan: Predicting hidden
  information in the starcraft fog of war with generative adversarial nets,''
  in \emph{Proceedings of the AAAI Conference on Artificial Intelligence},
  vol.~34, no.~04, 2020, pp. 4296--4303.

\bibitem{TemporalPoint}
E.~Merrill, S.~Lee, L.~Fuxin, T.~G. Dietterich, and A.~Fern, ``Deep convolution
  for irregularly sampled temporal point clouds,'' in \emph{arXiv preprint
  arXiv:2105.00137}, 2021.

\bibitem{SampleEfficientHebbian}
G.~Lagani, F.~Falchi, C.~Gennaro, and G.~Amato, ``Hebbian semi-supervised
  learning in a sample efficiency setting,'' in \emph{arXiv preprint
  arXiv:2103.09002}, 2021.

\bibitem{CompetingHebbian}
D.~Krotov and J.~J. Hopfield, ``Unsupervised learning by competing hidden
  units,'' \emph{Proceedings of the National Academy of Sciences}, vol. 116,
  no.~16, pp. 7723--7731, 2019.

\bibitem{HebbNet}
M.~Gupta, A.~Ambikapathi, and S.~Ramasamy, ``Hebbnet: A simplified hebbian
  learning framework to do biologically plausible learning,'' in \emph{ICASSP
  2021-2021 IEEE International Conference on Acoustics, Speech and Signal
  Processing (ICASSP)}.\hskip 1em plus 0.5em minus 0.4em\relax IEEE, 2021, pp.
  3115--3119.

\bibitem{MultiLayerHebbian}
T.~Miconi, ``Multi-layer hebbian networks with modern deep learning
  frameworks,'' in \emph{arXiv preprint arXiv:2107.01729}, 2021.

\bibitem{HebbianCluster}
X.~Hu, J.~Zhang, P.~Qi, and B.~Zhang, ``Modeling response properties of v2
  neurons using a hierarchical k-means model,'' \emph{Neurocomputing}, vol.
  134, pp. 198--205, 2014.

\bibitem{PointNet}
C.~R. Qi, H.~Su, K.~Mo, and L.~J. Guibas, ``Pointnet: Deep learning on point
  sets for 3d classification and segmentation,'' in \emph{Proceedings of the
  IEEE conference on computer vision and pattern recognition}.\hskip 1em plus
  0.5em minus 0.4em\relax IEEE, 2017, pp. 652--660.

\bibitem{InstarRule}
S.~Grossberg, ``Adaptive pattern classification and universal recoding: I.
  parallel development and coding of neural feature detectors,''
  \emph{Biological cybernetics}, vol.~23, no.~3, pp. 121--134, 1976.

\bibitem{ConvLSTM}
X.~Shi, Z.~Chen, H.~Wang, D.~Y. Yeung, W.~K. Wong, and W.~C. Woo,
  ``Convolutional lstm network: A machine learning approach for precipitation
  nowcasting,'' in \emph{Advances in neural information processing systems 28},
  2015.

\bibitem{HSNN}
X.~She, S.~Dash, D.~Kim, and S.~Mukhopadhyay, ``A heterogeneous spiking neural
  network for unsupervised learning of spatiotemporal patterns,''
  \emph{Frontiers in Neuroscience}, vol.~14, p. 1406, 2021.

\end{thebibliography}
\end{document}